\documentclass[letterpaper,twocolumn,10pt]{article}
\usepackage{usenix,epsfig,endnotes}
\usepackage{xcolor}
\usepackage{titleps}
\usepackage[letterpaper, margin=0.95in]{geometry}
\usepackage{url}
\usepackage{amsmath}
\usepackage{amssymb}
\usepackage{wrapfig}
\usepackage{float}
\usepackage{mathtools}
\usepackage{enumitem}
\usepackage{tabu}
\usepackage{parskip}
\usepackage{natbib}
\usepackage{amsmath}
\usepackage{bbm}
\usepackage{comment}
\usepackage{listings}
\usepackage{graphicx}
\usepackage{subcaption}
\begin{document}

\date{}

\title{\Large \bf Can Transformers Learn Sequential Function Classes In Context?
}

\author{
{\rm Ryan Campbell\endnote{Contributions from Ryan Campbell: thought of and formalized the notion of the sequential sliding window function class; wrote code to run the recursive two-layer ReLU neural networks; helped debug failed experiments such as initial numerical stability issues, unreasonable failures on simple tasks, unrealistic success on extremely difficult tasks, etc.; ran many of the experiments with my RTX 4090; wrote much of sections 3, 4, and 5}} \\
ryancampbell@berkeley.edu
\and
{\rm Emma Guo\endnote{Contributions from Emma Guo: Debugged the base linear sequential model (contributed to the design of the L2-normalized loss), designed and wrote the code for the randomization experiments, ran and debugged many experiments with sequential NN and linear sequential models.}} \\
emmaguo@berkeley.edu
\and
{\rm Evan Hu\endnote{Contributions from Evan Hu: came up with and led the initial direction of the project. Developed the base linear sequential function task and worked on training the model. Formalized and wrote up many of the findings from the experiments into the paper. }} \\
evanhu@berkeley.edu
\and
{\rm Reya Vir\endnote{Contributions from Reya Vir: worked on an approach regarding the rethinking demonstrations paper, before we decided to focus on another approach. Helped with debugging for sequential and ReLU functions, and helped with debugging randomized labels code. contributed to paper writing process and wrote self-review. addressed revisions in the paper and code, and wrote response to reviewers. }} \\
reyavir@berkeley.edu
\and
{\rm Ethan Hsiao} \\
ethan.hsiao@berkeley.edu
}
\maketitle

\thispagestyle{empty}

\subsection*{Abstract}

In-context learning (ICL) has revolutionized the capabilities of transformer models in NLP. In our project, we extend the understanding of the mechanisms underpinning ICL by exploring whether transformers can learn from sequential, non-textual function class data distributions. We introduce a novel sliding window sequential function class and employ toy-sized transformers with a GPT-2 architecture to conduct our experiments. Our analysis indicates that these models can indeed leverage ICL when trained on non-textual sequential function classes. Additionally, our experiments with randomized y-label sequences highlights that transformers retain some ICL capabilities even when the label associations are obfuscated. We provide evidence that transformers can reason with and understand sequentiality encoded within function classes, as reflected by the effective learning of our proposed tasks. Our results also show that the performance deteriorated with increasing randomness in the labels, though not to the extent one might expect, implying a potential robustness of learned sequentiality against label noise. Future research may want to look into how previous explanations of transformers, such as induction heads and task vectors, relate to sequentiality in ICL in these toy examples. Our investigation\endnote{The GitHub can be found here: \url{https://github.com/emmaguo13/in-context-learning}} lays the groundwork for further research into how transformers process and perceive sequential data.


\section{Introduction}

In Context Learning (ICL) in transformers represents a pivotal development in the field of natural language processing (NLP), allowing a paradigm shift in how machine learning models understand and generate language. This approach, exemplified by models like GPT-3 and GPT-4, relies on the model's ability to learn from the context provided within the prompt itself, without the need for explicit fine-tuning or retraining. In-context learning, first discussed in-depth in Brown et al.'s groundbreaking paper "Language Models are Few-Shot Learners" \cite{brown2020fewshotlearners}, leverages the vast knowledge encoded in the parameters of large-scale language models. The concept further evolved with subsequent research \cite{gardner2020decisionboundaries}, which explored the nuanced understanding that transformers demonstrate when presented with slightly altered inputs. While much progress has been made in understanding the effectiveness of ICL and how to leverage it through effective prompt engineering \cite{wei2022chainofthought}, there is still little known about the mechanisms within neural networks that enable ICL.

To this end, many recent bodies of literature have attempted to explore the relationship between model architecture, training data distribution, and other intrinsic properties of a meta learning model in producing emergent ICL behavior. 

Our project focuses on investigating the role that the sequentiality of the training data plays in producing models capable of ICL. Specifically, we investigate whether or not Transformers are able to in-context learn sequential non-textual data, defined formally below in section 3.2. We build upon the paper “What Can Transformers Learn In-Context? A Case Study of Simple Function Classes” \cite{garg2023transformers} and extend the analysis of Transformers and simple function classes with sequential data distributions and a novel sliding window sequential function class. 

We also re-implement the experiment from “Rethinking the Role of Demonstrations: What Makes In-Context Learning Work?” \cite{min2022rethinking} and randomly permute the labels for our sequential data inputs to examine if ICL ability is still preserved with our sequential non-textual function classes. In contrast to the findings from the paper, where ICL performance was preserved despite randomized / irrelevant in context pairs (i.e. $(x_i, y_j)_{i\neq j}$), we actually find that ICL performance degrades with non-textual sequential data. We hypothesize that this occurs because some of the fundamental mechanisms underlying ICL are different for textual and non-textual meta learning tasks involving sequential data and propose several deeper analyses/explanations further in the Findings section.

Our contributions are as follows:

\begin{enumerate}
  \item We defined a novel non-textual sequential sliding window function class.
  \item We trained toy-sized GPT2-architecture transformers on these new sequential tasks and found that transformers are capable of ICL when trained on non-textual sequential function classes.
  \item We evaluated the same tasks with randomized labels for our sequences and found that this degrades ICL performance in transformers, suggesting that transformers trained on non-textual sequential meta-learning tasks are less robust than with textual tasks.
  \item Based on these two results, we claim that transformers are able to learn and understand sequentiality for non-textual function class data distributions and hypothesize that there are fundamentally different mechanisms underlying ICL  for textual and non-textual meta-learning sequential tasks.
\end{enumerate}

\section{Related Works}

There have been numerous explanations proposed for the emergence of ICL (in-context learning), but many of these are in tension with one another. 

In \cite{olsson2022incontext}, the authors observe that a significant part of in-context learning can be explained by specific attention heads, which they call induction heads. Specifically, these arise from pairs of attention heads which work together - the first head copies information from the previous token into each token, and the second head, the induction head, searches in the context of the first head. This is seen especially in smaller models, and in our project, we were interested in investigating if these induction heads appear in sequential data as well, where the order of the input is relevant for the output.

Another paper, \cite{garg2023transformers}, proposes a similar idea, that pattern matching accounts for a majority of in-context learning, and they also observe that adding noise to the labels hurts performance, which we later show aligns with our experiments. However, in \cite{min2022rethinking}, the authors found that when the model is provided with random “y” labels, the accuracy for in context (x, y) performs similarly to when the model is provided with accurate (“gold”) labels. This begs the question of whether the labels provided may not matter as much as we previously expected, and the model is primarily focusing on learning the structure of the input and output. We assert that this may be a result of language being a more unstructured task, where learning structure plays a large role in accuracy.

We believe the differing conclusions stem from key differences between the tasks studied. Language modeling involves more complex and unstructured data, so structural cues could play a bigger role. However, we are training on simple toy function classes rather than language data. The paper also uses much larger pretrained models, specifically GPT2-large, which could be the reason for higher accuracies on their test data. Given these observations, we decided that while this paper provides interesting insights on the capabilities of transformers for in-context learning, the results may not give much insight on our study on toy tasks.

\section{Formal Background}

\subsection{In-context Learning}

The following formalization of ICL is given by \cite{garg2023transformers}:

In order to in-context learn some function class $\mathcal{F}$, sample some $f$ from a distribution over functions, $\mathcal{D}_{\mathcal{F}}$, and sample a prompt sequence $(x_1, f(x_1),\ldots,x_k,f(x_k),x_\text{query})$ from a distribution over inputs, $\mathcal{D}_{\mathcal{X}}$. Then, a model $M$, in this case a transformer, in-context learns a function class $\mathcal{F}$ up to an error $\epsilon$ if 
$$\mathbb{E}_P\left[l(M(P),f(x_\text{query}))\right]\leq\epsilon,$$ 
where $P=(x_1, f(x_1),\ldots,x_k,f(x_k),x_\text{query})$ denotes a prompt.

\subsection{Sequential Function Classes}

While \cite{garg2023transformers} investigates function classes that are order-invariant (linear regression, two-layer neural networks, decision trees, etc.), this paper departs from this and investigates a novel characterization of sequential toy function classes. \\

For our sequential function classes, we sample an initial $x_{0}$ from a Gaussian distribution and define a recursive relation that is recursively applied to obtain a sequence of length $n$. When training, the label of each input is simply the value of the next input in the sequence (i.e. our Y label sequence is just the X input sequence shifted by one). We introduce a sliding window / convolution to the recursive relation such the previous $k$ where $k\in[1,n]$ items in the sequence are aggregated to get the value of the next input. Formally:

$$x_0 \sim\mathcal{N}(0, I)$$
$$x_{i+1}=F(x_i,x_{i-1},\ldots,x_{i-k})=\sum_{j=0}^kf_j(x_{i-j})$$
$$y_i=x_{i+1}$$
Note that for the first $k-1$ inputs, we used a default value of $0$ for $x_{\leq0}$. \\

We chose this definition because this is analogous to the sequential relationship language tokens have with the previous tokens in the sequence.
The above definition implies that the shape of the inputs and the outputs will be the same. It also assumes, which our experiments validate, that the sequential function is linearly separable with respect to the inputs $x_i,\ldots,x_{i-k}$. \\

\subsection{Numerical Stability}

Since these sequential functions are recursively defined, it will be important to consider the numerical stability of the functions $F$ sampled from $\mathcal{D}_\mathcal{F}$. \\

For example, consider a simple sequential function class defined by the following:
$$\left\{y=Wx:W\in\mathbb{R}^{n\times n}\right\}.$$
Then, the numerical stability of such sequences is determined by the magnitude of the eigenvalues as well as the length of the sequence. For shorter sequences, anything should be fine, but it may not perform as well due to the lack of in-context examples. With longer sequences, if there are eigenvalues with magnitude greater than $1$, the inputs and outputs will explode in size. This can be seen by considering the singular value decomposition (SVD):
$$W^ix_i=(U\Sigma V^\mathsf{T})^ix_i=U\Sigma^iV^\mathsf{T}x_i$$
For eigenvalues greater than $1$ and large $i$, the values of the diagonal matrix $\Sigma^i$ will be exponentially large. Thus, generating matrices with restricted eigenvalues ($|\lambda|\leq 0.99$) will keep the values within reason.

\section{Experimental Setup}

Much of the code used to run our experiments is built on top of the repository from \cite{garg2023transformers}. In order to experiment with our novel definition of sequential functions, we defined a new task, the sliding window sequential task, which is a special case of the tasks defined before. From this, we further defined three special cases of sequential tasks:
\begin{enumerate}
    \item \emph{Recursive bias}:
    $$x_{i+1}=x_i+b_i$$
    This sequence uses a sliding window of length $1$, and only grows multiplicatively, which is fine for reasonable sequence lengths.
    \item \emph{Recursive linear transformations}: 
    $$x_{i+1}=Wx_i$$
    This sequence also uses a sliding window of length $1$, and stays numerically stable by limiting eigenvalues.
    \item \emph{Recursive two-layer ReLU neural networks with normalization}:
    $$x_{i+1}=\frac{\sum_{j=0}^kf_j(x_{i-j})}{\|\sum_{j=0}^kf_j(x_{i-j})\|_2},$$
    where $f_j$ are neural networks. This uses a sliding window of length $k$. Since there is normalization at each step, the sequence is numerically stable. 
\end{enumerate}

For the first two sequential tasks, we also modified the loss function to account for numerical stability issues. We used the following loss function:
$$l(y,\hat{y})=\frac{\left\|y-\hat{y}\right\|_2}{\left\|y\right\|_2}$$
By normalizing with respect to $y$, the correctness of larger and small values of $y$ is emphasized equally in the gradients. Without this scaling factor, the loss was dominated by the later terms in the sequence when the inputs and outputs grew larger. Note that it was not necessary to do this for the third task since normalization occurred at every step in the sequence anyways. \\

Another departure from \cite{garg2023transformers} is that instead of sampling the entire prompt from $\mathcal{D}_\mathcal{X}$, only the initial input is sampled, and the rest is generated by repeatedly applying $F$. \\

During training, each $x_i$ in the prompt is treated as an $x_\text{query}$ and some point, and then the loss is used to update the model. When each $x_i$ is used, the model only has access to the embeddings of the previous $(x_j,y_j)$ pairs. \\

We trained each model with a different set of dimensions and examples. 
\begin{table}[h]
\centering
\begin{tabular}{|c|c|c|} 
 \hline
 & Dimensions & Examples \\ \hline
recursive bias & 20 & 20 \\ \hline  
recursive linear & 20 & 11 \\ \hline
recursive nn & 20 & 100 \\ \hline
\end{tabular}
\caption{Model Training Parameters}
\label{tab:mytable}
\end{table} 

Our GPT2 model had 12 layers, 8 attention heads, and embedding sizes of 256.

\subsection{Randomization of Labels}

In order to analyze the robustness of training using sequences of in context examples, we observe how the model responds to noise by randomizing select $y$ values in the context sequence used for evaluation. \\

Our analysis involved randomizing $r$ selected sequence outputs, where $r$ goes from $1$ to one less than the maximum sequence length, and seeing how the randomized outputs effect squared loss the more in-context examples we have. We then repeat this analysis by randomizing both the inputs and the outputs.

In order to generate the random input/output replacements, we sample random values from a Gaussian distribution to replace the outputs within the context. To evaluate, we then pass the partly randomized contexts into the model's forward pass and evaluate the squared loss on the query output. \\

Note that in the case where we only randomize the outputs in the sequence, a sequential model still has access to the same $y$ values that were randomly reset, as they are stored in the $x$ values in the prompt. Thus, it is still feasible for the model to perform well even after this perturbation. However, in the case where we randomize $r$ inputs and outputs, the model will not be able to get any useful context from the randomized portion of the sequence. 

\section{Findings}

Our experiments involved training transformers, specifically using a GPT2-like architecture, on non-textual sequential meta-learning tasks. We find that transformers are able to learn non-textual sequential function tasks in context. 

We also discovered that when the y-labels of the training sequences were randomly permuted, the performance of the transformers in In-Context Learning (ICL) was notably degraded. This outcome indicates a clear difference in how transformers handle textual versus non-textual sequential tasks, suggesting that transformers, while capable of learning sequence patterns in non-textual data, exhibit less robustness in these scenarios compared to textual tasks.

We further observed that transformers are indeed capable of understanding and learning from sequential non-textual function class data distributions. However, the performance varied significantly when subjected to label randomization. In tests involving recursive two-layer ReLU neural networks with normalization, transformers demonstrated a decline in performance when evaluated with randomized sequence labels.

Though our findings were based on experiments on toy models, it reveals significant properties of transformers and their capabilities to learn sequential patterns beyond natural language.

\subsection{Recursive Bias}

\begin{figure}[h]
    \centering
    \includegraphics[width=0.8\linewidth]{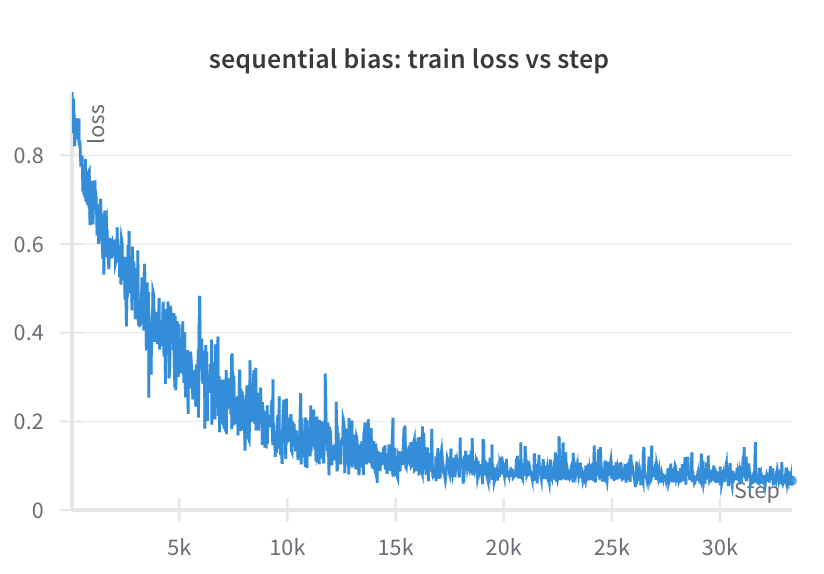}
    \caption{Train loss for the recursive bias task}
    \label{fig:recursive_bias}
\end{figure}

As seen in figure \ref{fig:recursive_bias}, the GPT2 architecture was able to perform well on the recursive bias task. \\

However, it should be noted that the loss did not get as close to $0$ as expected. This may be due to the fact that this task was the least numerically stable of all of the tasks. Since a bias is added at each step, the size of the values depend mostly on the sequence length. 

\subsection{Recursive Linear Transformations}

\begin{figure}[h]
    \centering
    \includegraphics[width=0.8\linewidth]{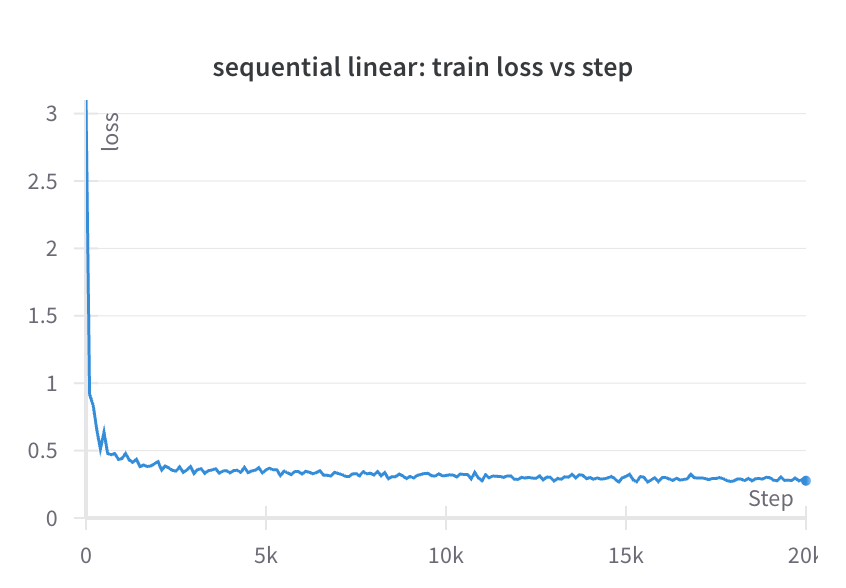}
    \caption{Train loss for the recursive linear transformation task}
    \label{fig:recursive_linear}
\end{figure}

As seen in figure \ref{fig:recursive_linear}, the GPT2 architecture was able to perform well on the recursive linear transformation task. \\

\subsection{Recursive Two-layer ReLU Neural Networks with Normalization}

\begin{figure}[h]
    \centering
    \includegraphics[width=0.8\linewidth]{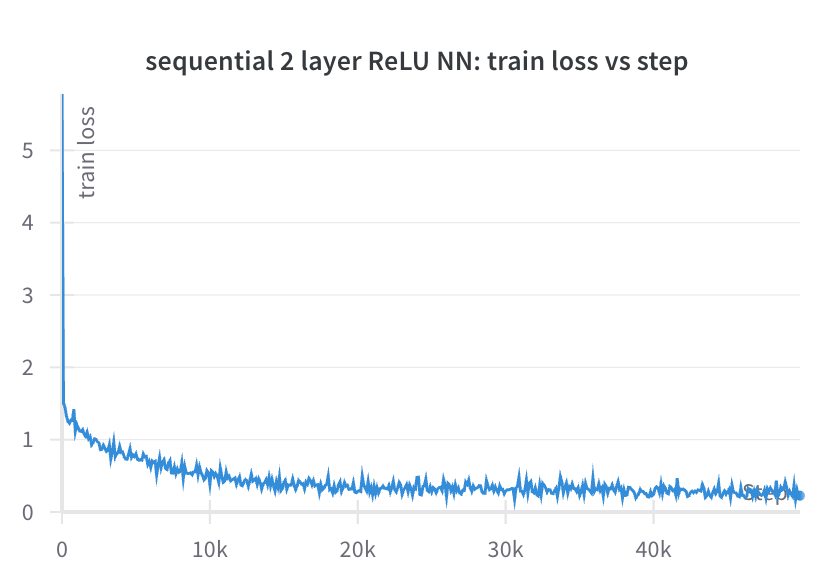}
    \caption{Train loss for the two-layer ReLU neural networks with normalization task}
    \label{fig:recursive_relu}
\end{figure}

As seen in figure \ref{fig:recursive_relu}, the GPT2 architecture was able to perform well on the recursive two-layer relu neural networks with normalization task. 

\begin{figure}
    \centering
    \begin{subfigure}{0.35\textwidth}
        \includegraphics[width=\linewidth]{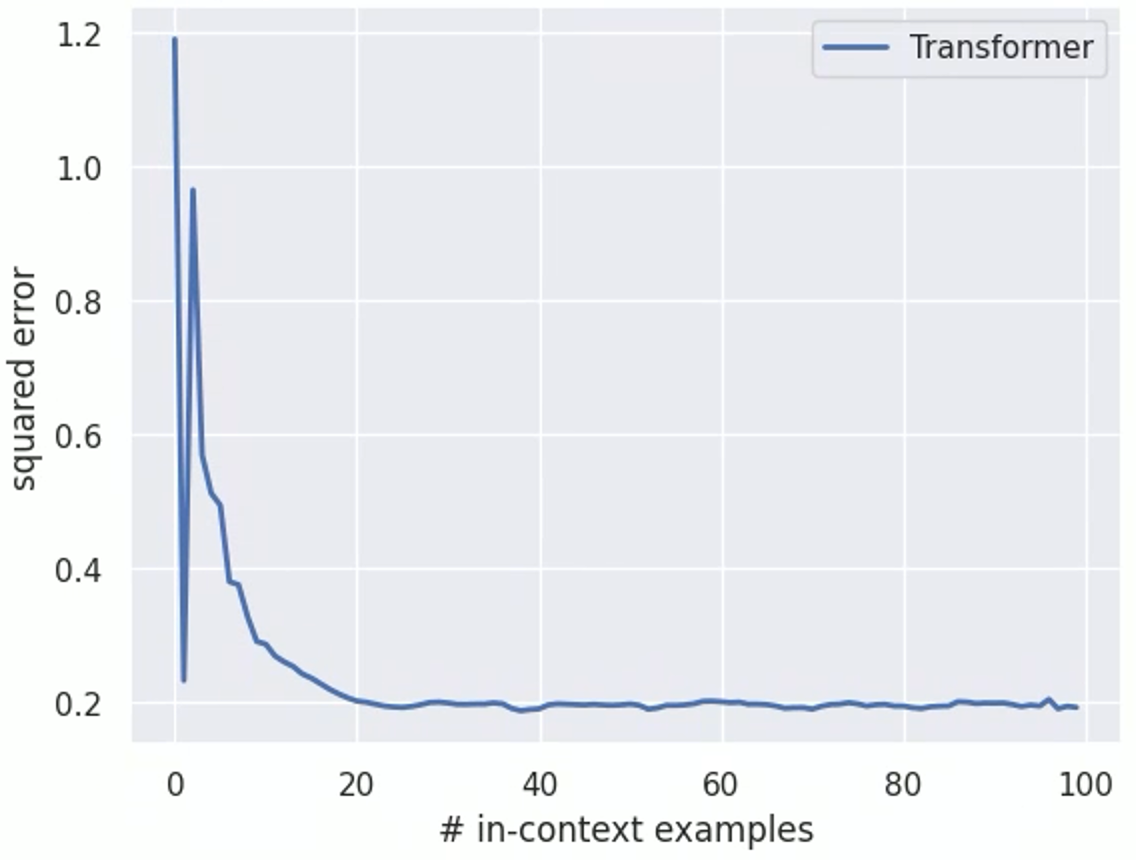}
        \caption{$0$ context outputs randomized}
        \label{fig:sub1}
    \end{subfigure}
    \vspace{1em}
    \begin{subfigure}{0.35\textwidth}
        \includegraphics[width=\linewidth]{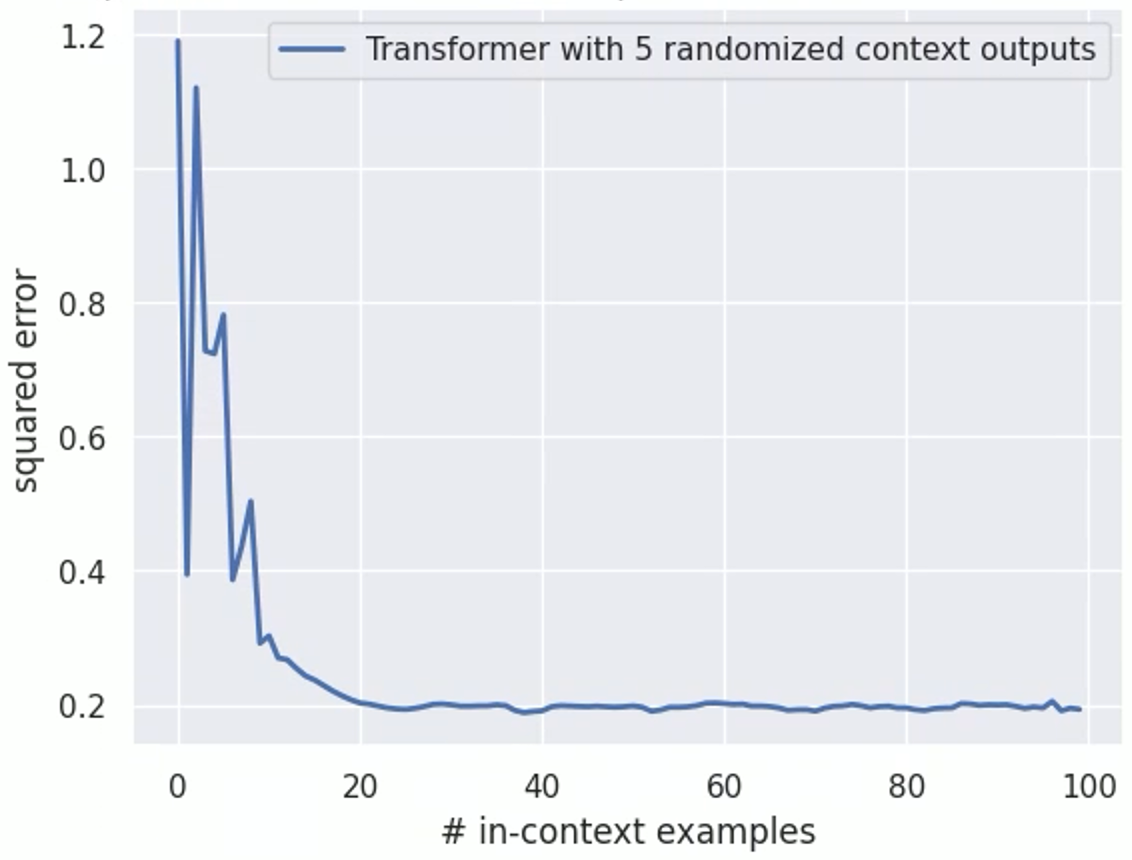}
        \caption{$5$ context outputs randomized}
        \label{fig:sub2}
    \end{subfigure}
    \vspace{1em}
    \begin{subfigure}{0.35\textwidth}
        \includegraphics[width=\linewidth]{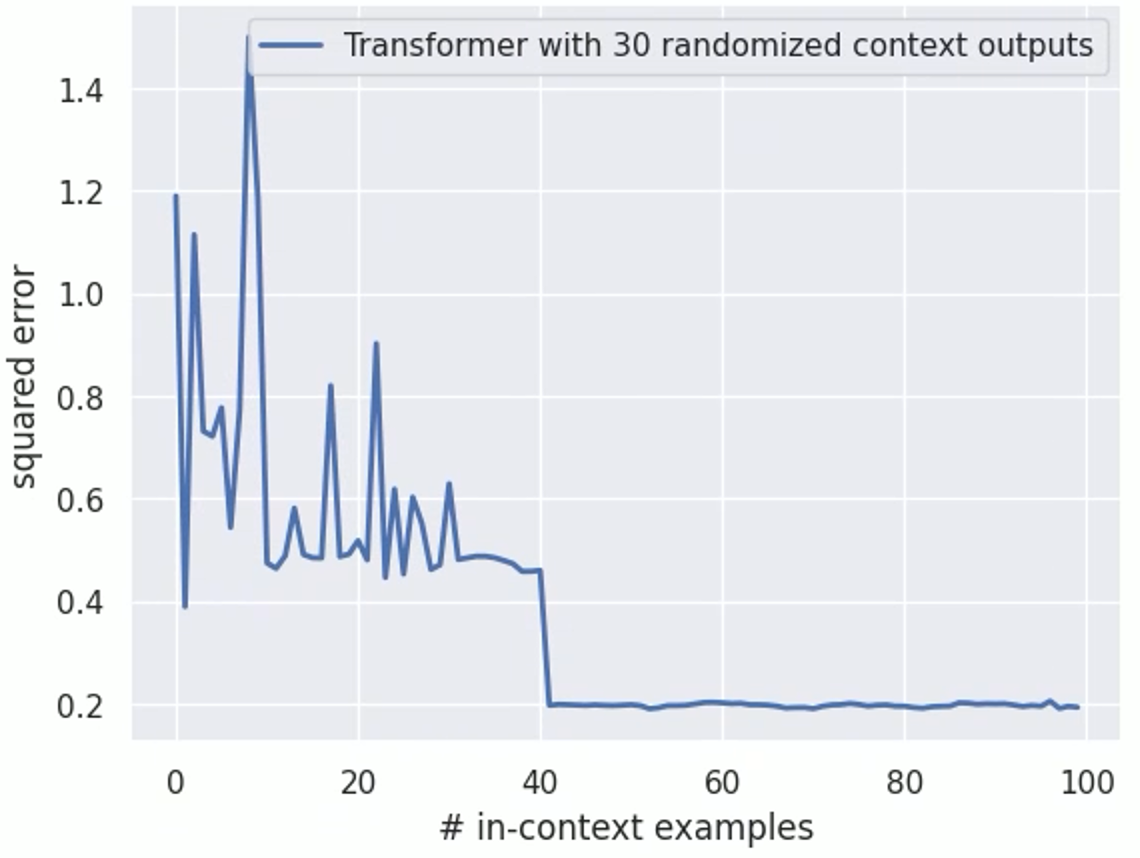}
        \caption{$30$ context outputs randomized}
        \label{fig:sub3}
    \end{subfigure}
    \vspace{1em}
    \begin{subfigure}{0.35\textwidth}
        \includegraphics[width=\linewidth]{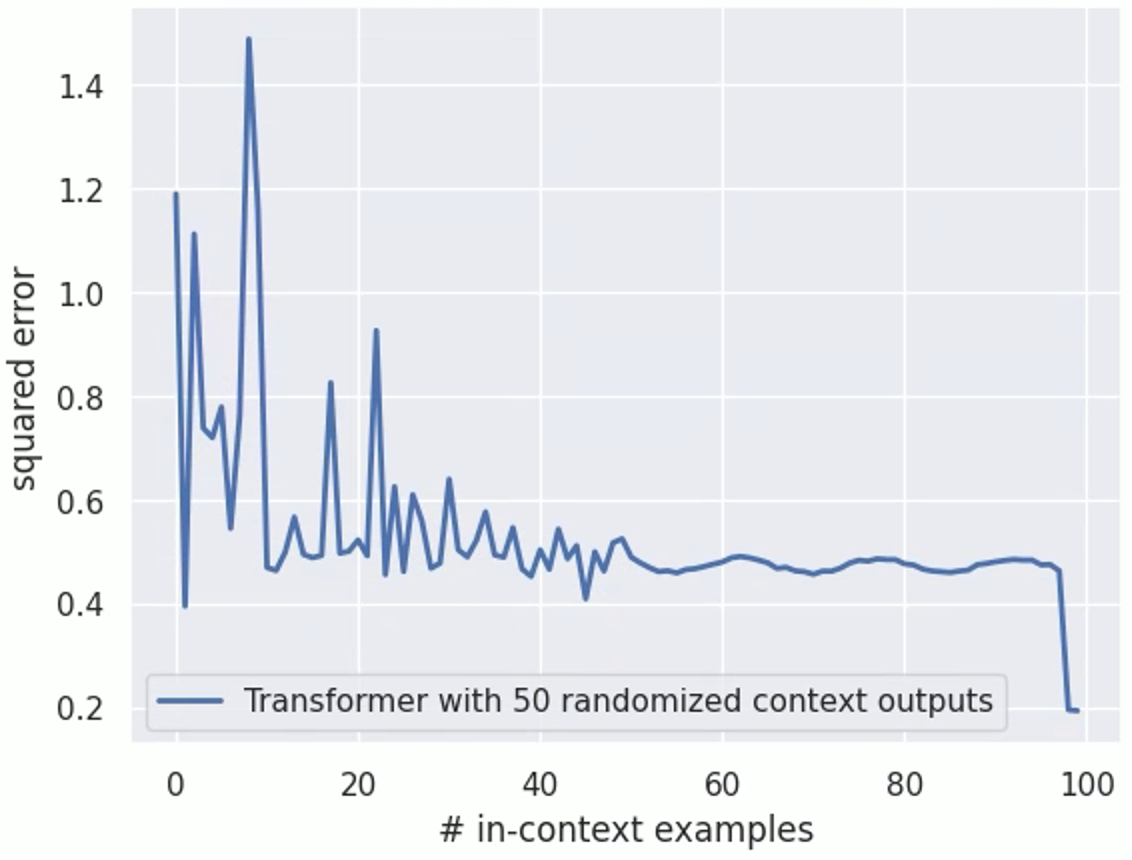}
        \caption{$50$ context outputs randomized}
        \label{fig:sub4}
    \end{subfigure}
    \caption{Squared error vs the number of in-context examples for the model trained on the recursive two-layer ReLU neural network, with variable number of outputs randomized}
    \label{fig:main}
\end{figure}

Additionally, the same model was evaluated on prompts with randomized context outputs in order to test its robustness. As shown in figure \ref{fig:main}, increasing the number of randomized context outputs makes it harder for the model to successfully predict the next output. That is, with a worse context, the model performs worse. This suggests that the model was in fact learning from the context and not just making general assumptions about the averages in the data. 

\subsection{Randomization of Labels}

Finally, we reran our recursive two-layer ReLU neural networks with normalization experiments except we inject noise by randomly permuting the labels of in-context pairs in the prompt, finding that unlike the result in \cite{min2022rethinking},  ICL performance degrades with non-textual sequential data.

We plotted the squared loss of the query output given a context with $r$ randomized $y$ values\endnote{See the appendix for more examples of randomized context outputs.}. We notice that when $r$ is less than half of the total in context examples, in-context learning still occurs. This supports the robustness of the model after being trained for over 5k steps with sequences containing contextual examples of linear functions. After having trained on so many inputs with linear examples, the model is able to ignore significant amounts of noise, and interpolate the weights using the remaining examples with linear structure.

It's possible that the structure of the meta learning in-context prompt itself 
is what makes the difference. In textual tasks, the in-context prompt often contains a sequence of examples that are directly related to the text being processed, allowing the model to establish patterns and relationships within the context of the text as well as leverage understanding gained from the pre-training. However, in non-textual sequential tasks, this direct relationship between the examples in the in-context prompt and the task at hand may not be as clear or as easily discernible.

This could lead to a situation where the in-context learning (ICL) mechanisms that work well for text-based tasks are less effective for non-textual sequential data. The sequential nature of non-textual data may require a different approach to pattern recognition and learning. For example, in tasks involving images, time-series data, or other non-textual sequences, the importance of spatial or temporal relationships, feature extraction, and the representation of these features might play a more significant role than in text-based tasks.

Therefore, the degradation in ICL performance with non-textual sequential data could stem from a mismatch between the ICL mechanisms optimized for text and the unique requirements of non-textual sequential tasks. This suggests that further research is needed to develop and refine ICL approaches specifically tailored for non-textual data, taking into account the distinct characteristics and learning dynamics of these types of data. Such research could lead to more robust and effective meta-learning models that can handle a wider variety of tasks, including those involving complex sequential non-textual data.

\section{Limitations and Next Steps}

Our compute power was limited. We trained on small transformers on the order of $\approx10$ million parameters on toy function classes. We had access to an RTX 4090, which we used for most of our training. In \cite{garg2023transformers}, the  authors state that they “train using a single NVIDIA GeForce RTX 3090 GPU and most training runs take 5-20 hours depending on model size and context length.” Our available compute was sufficient for the scale of experimentation that we conducted, but we were limited in further and more extensive testing. As a next step, we would like to conduct a deeper investigation into sequentiality with other sources of ICL proposed in recent literature, specifically induction heads.

Furthermore, we would have liked to investigate more deeply into induction heads as another approach to examining the role of sequentiality of data in the emergence of ICL in transformers. Specifically, to improve the paper, we would also extend` the work done by \cite{olsson2022incontext} in order to investigate whether induction heads also emerge when training on our non-textual sequential function class data. Furthermore, if as they claim, that induction heads are responsible for the “majority” of ICL behavior, then we are interested in understanding whether or not induction heads also appear when training on strictly non-sequential function class data as done in  \cite{garg2023transformers}.

We outline three critical directions for future investigation:

\begin{enumerate}
  \item \textbf{Visualization of Induction Heads:}
  A key objective is to explore the visualization of induction heads, particularly in relation to the simple sequential function classes we propose in this paper. The rationale is that if induction heads are crucial in sequential text ICL, then they should also manifest in our non-textual sequential function classes. This exploration also seeks to determine whether the presence or absence of induction heads is influenced by the complexity of the function. 
  \item \textbf{In-Context Learning with Unstructured Data:}
  The project also aims to delve into tasks involving unstructured data. Initially, the focus was on complex examples like regular expressions and natural language, as opposed to elementary function classes. Given additional computational resources, the research would expand to discern how in-context learning differs between structured and unstructured data. This exploration is inspired by the observation in the "Rethinking demonstrations" paper, where learning seemed equally effective with random versus gold labels, possibly due to the inherently complex and unstructured nature of language data. Understanding the extent to which learning the structure of data contributes to in-context learning is a key goal.
  \item \textbf{Exploration of Task Vectors:}
  The concept of "Task Vectors" as described in the paper suggests that these are primary drivers of implicit in-context learning, linking contextual training data to a "rule application" function. The next phase involves a deeper investigation into task vectors to ascertain whether they, along with the learning algorithm and rule function, are fundamental to in-context learning. This research will also examine the interplay between task vectors and induction heads, given their apparent significant roles in facilitating in-context learning.

\end{enumerate}

Another limitation is that the sequential functions were defined to be linearly separable with respect to the inputs. In the future, defining more general sequential functions would be an interesting direction as this would be make the toy functions even more general, and thus better to extract a deeper understanding from. For example, note that in language, the previous tokens are not even close to linearly separable when predicting the next token.

\newpage

{\footnotesize 
\bibliography{sample}}


\section*{Appendix}

\begin{figure}[h]
    \centering
    \includegraphics[width=0.8\linewidth]{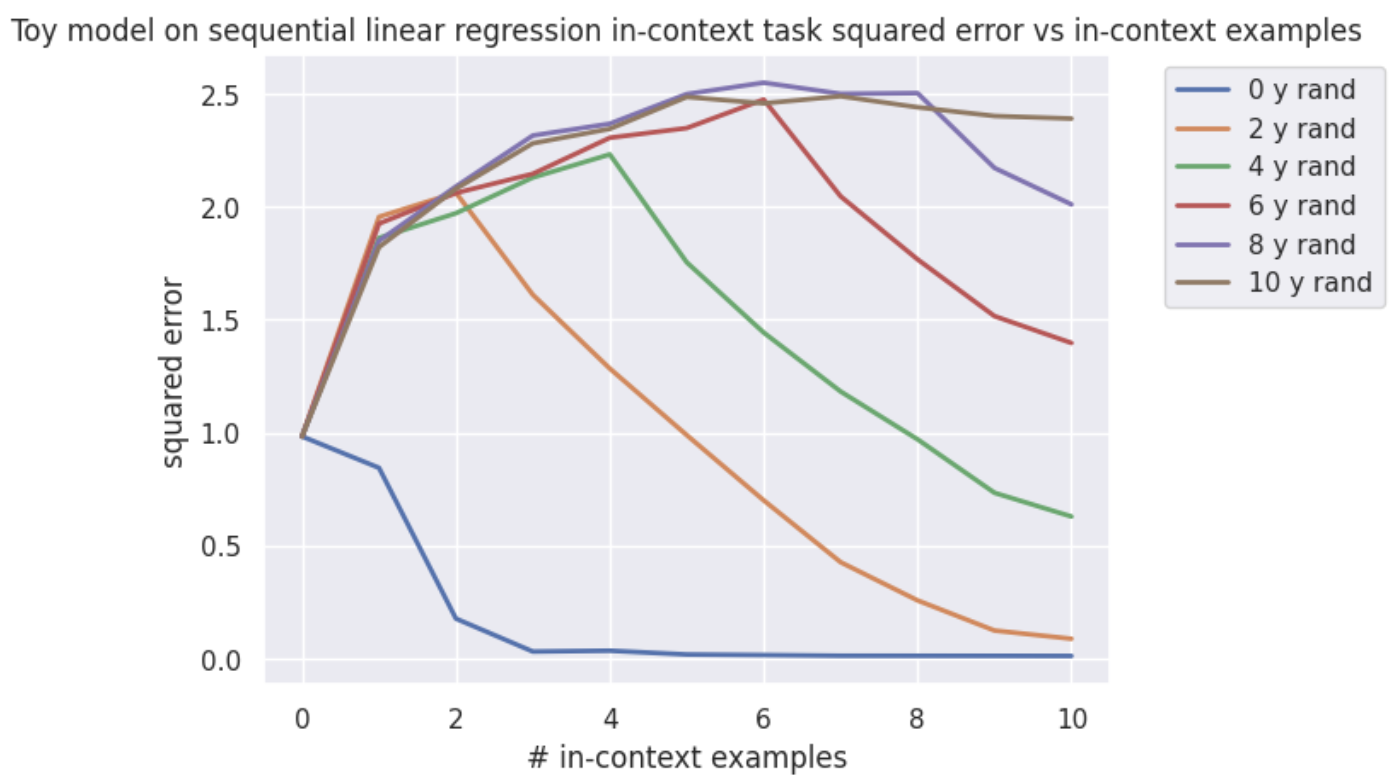}
    \caption{The toy sequential linear model's evaluation loss over the number of in-context examples given, where y values in the context are randomized.}
    \label{fig:5}
\end{figure}

\begin{figure}
    \centering
    \includegraphics[width=0.8\linewidth]{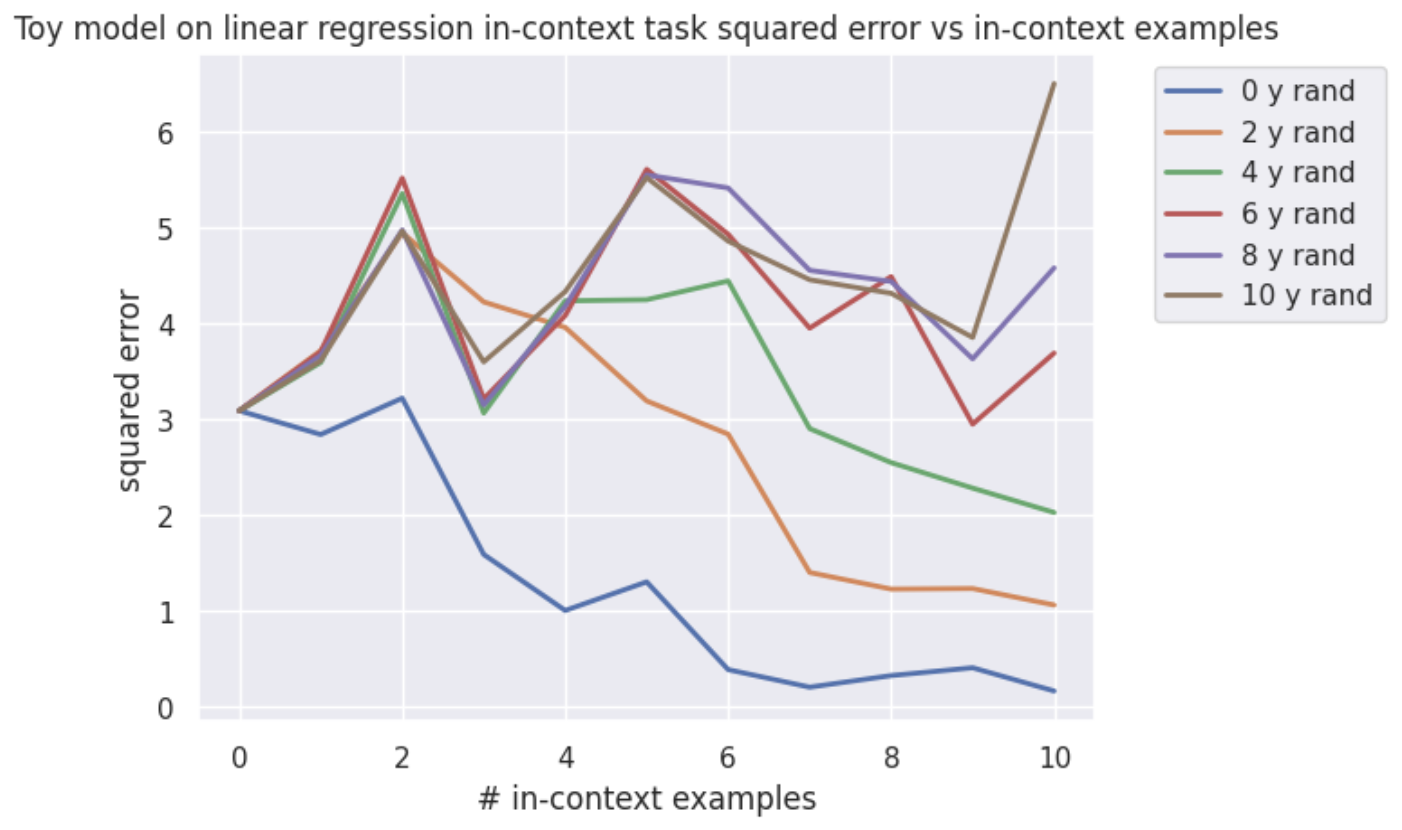}
    \caption{The toy linear model's evaluation loss over the number of in-context examples given, where y values in the context are randomized.}
    \label{fig:6}
\end{figure}

\begin{figure}
    \centering
    \includegraphics[width=0.8\linewidth]{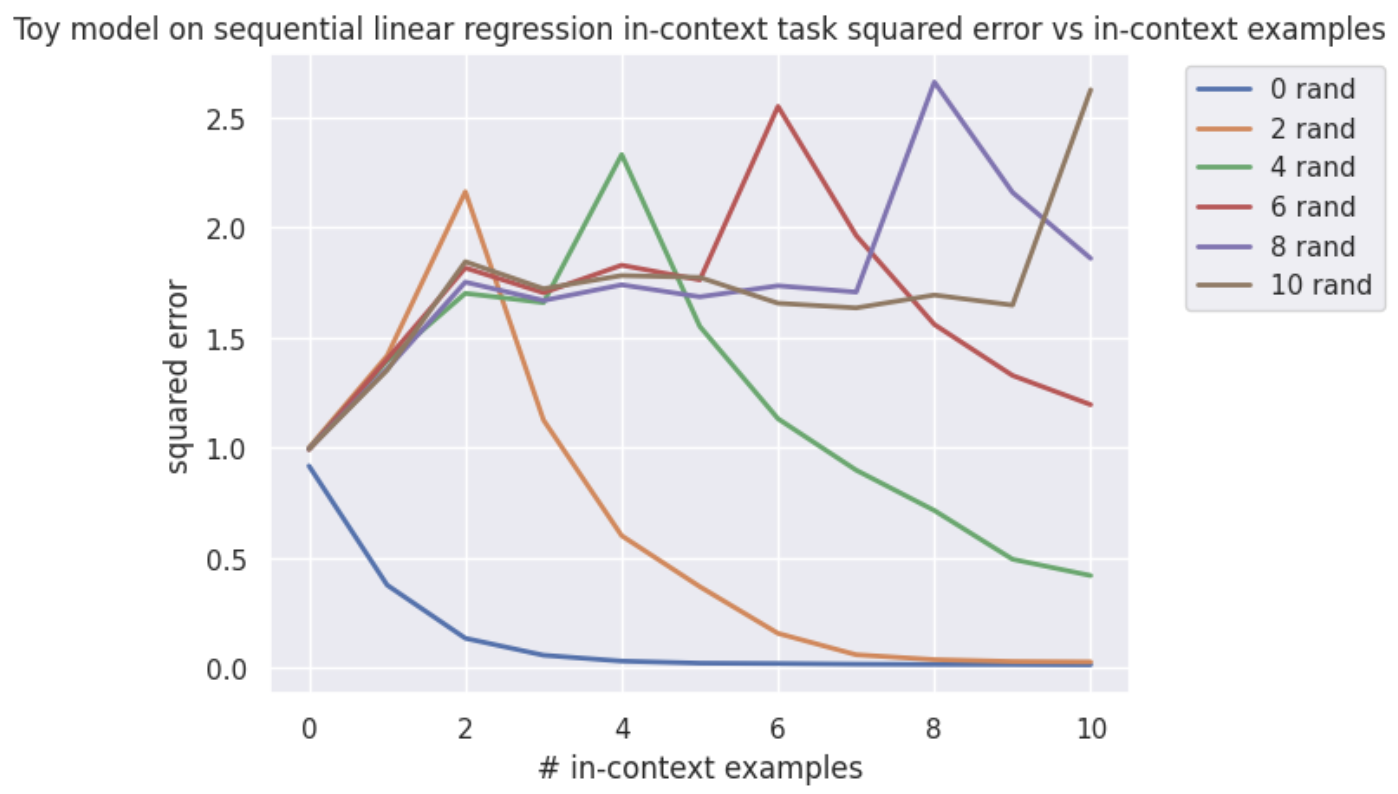}
    \caption{The toy sequential linear model's evaluation loss over the number of in-context examples given, where x and y values in the context are randomized.}
    \label{fig:7}
\end{figure}

\begin{figure}
    \centering
    \includegraphics[width=0.8\linewidth]{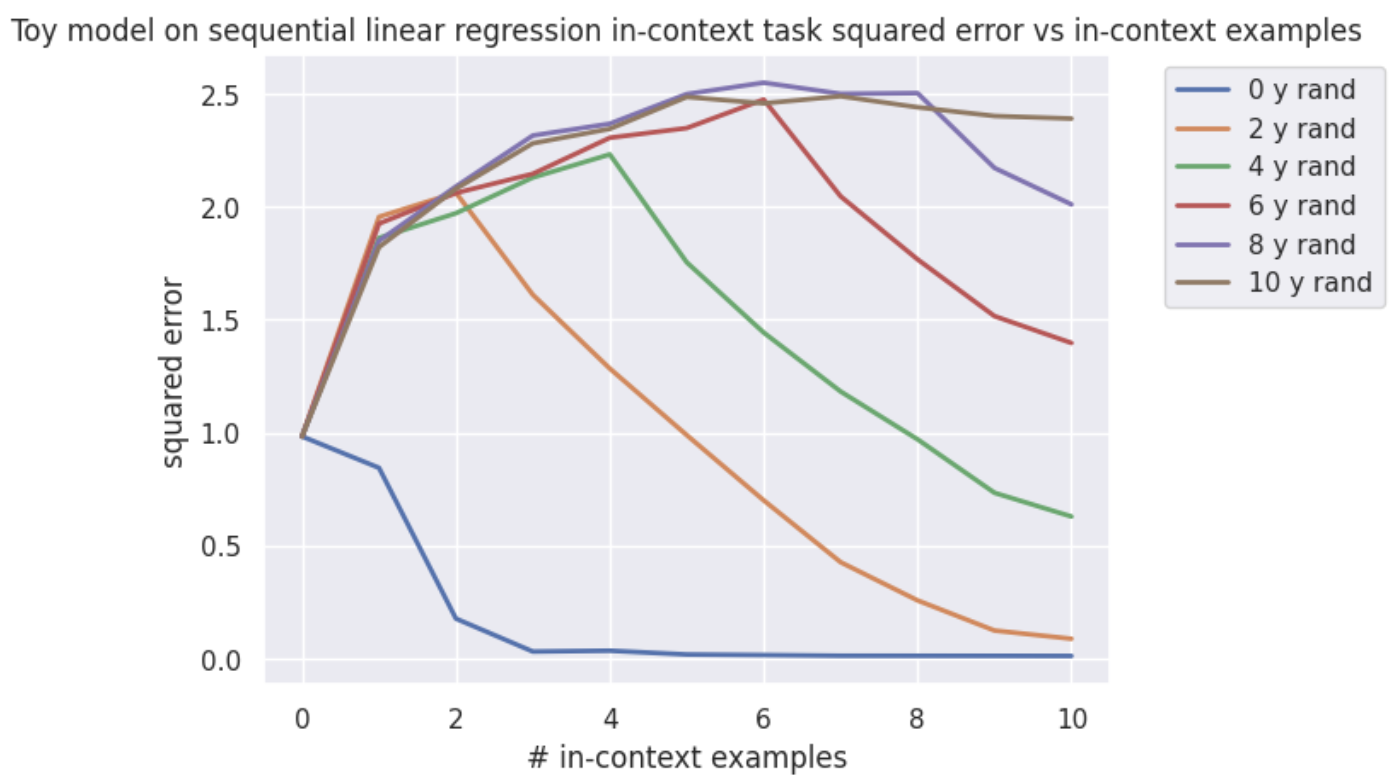}
    \caption{The toy linear model's evaluation loss over the number of in-context examples given, where x and y values in the context are randomized.}
    \label{fig:8}
\end{figure}

\newpage

\theendnotes

\end{document}